\documentclass[conference]{IEEEtran}
\IEEEoverridecommandlockouts
\usepackage{cite}
\usepackage{amsmath,amssymb,amsfonts}
\usepackage{graphicx}
\usepackage{textcomp}
\usepackage{xcolor}
\usepackage{booktabs}
\usepackage{framed}
\usepackage{adjustbox}
\usepackage{algorithm}%
\usepackage{algorithmicx}%
\usepackage{algpseudocode}%
\usepackage{listings}%
\usepackage{amssymb}
\usepackage{array}
\def\BibTeX{{\rm B\kern-.05em{\sc i\kern-.025em b}\kern-.08em
    T\kern-.1667em\lower.7ex\hbox{E}\kern-.125emX}}
\begin{document}

\title{Robust Neural Pruning with Gradient Sampling Optimization for Residual Neural Networks}

\newcommand{\JY}[1]{\textcolor{red}{[DANIEL: #1]}}
\author{\IEEEauthorblockN{Juyoung Yun}
\IEEEauthorblockA{\textit{Department of Computer Science, Stony Brook University} \\
New York, 11794, United States \\
juyoung.yun@stonybrook.edu}
}


\maketitle

\begin{abstract}
This research embarks on pioneering the integration of gradient sampling optimization techniques, particularly StochGradAdam, into the pruning process of neural networks. Our main objective is to address the significant challenge of maintaining accuracy in pruned neural models, critical in resource-constrained scenarios. Through extensive experimentation, we demonstrate that gradient sampling significantly preserves accuracy during and after the pruning process compared to traditional optimization methods. Our study highlights the pivotal role of gradient sampling in robust learning and maintaining crucial information post substantial model simplification. The results across CIFAR-10 datasets and residual neural architectures validate the versatility and effectiveness of our approach. This work presents a promising direction for developing efficient neural networks without compromising performance, even in environments with limited computational resources.
\end{abstract}

\begin{IEEEkeywords}
Neural Networks, Optimization, Neural Pruning
\end{IEEEkeywords}

\section{Introduction}
\label{Introduction}
The rapid evolution of neural networks has been crucial in advancing numerous applications across various fields~\cite{prun2021}. However, the increasing size and complexity of these models pose significant challenges, particularly in environments with constrained computational resources\cite{ding2023efficiency}. This has led to an intensified focus on network pruning, a technique essential for streamlining DNNs by removing redundant weights \cite{Cheng2017SurveyPruning}. Pruning not only enhances computational efficiency but also facilitates the deployment of neural networks in resource-limited settings~\cite{prun2019}.

Despite its advantages, a primary concern with pruning is the potential loss of accuracy~\cite{Cheng2017SurveyPruning}. This accuracy loss occurs because pruning can inadvertently remove weights that are crucial for the network's performance \cite{Han2015DeepCompression}. Addressing this challenge necessitates innovative approaches that can efficiently prune networks while preserving their accuracy.

We introduce a novel approach that integrates advanced gradient sampling optimization techniques, such as StochGradAdam~\cite{Yun2023StochGradAdam}, with the pruning process to maintain model accuracy despite significant parameter reductions. This method, characterized by selectively using a portion of the gradients and setting others to zero during the optimization step, offers a fresh perspective on neural network optimization. Unlike traditional methods like Adam~\cite{Kingma2014Adam}, where a substantial accuracy drop is often observed post-pruning, StochGradAdam shows great promise in preserving accuracy, providing a compelling alternative in the field.

Building on this foundation, our research delves into the efficacy of StochGradAdam in conjunction with a pruning technique across various architectures, including ResNet 56, 110, and 152~\cite{He2016ResNet}. These models, celebrated for their depth and complexity, serve as a robust platform for evaluating our approach’s effectiveness in maintaining accuracy during pruning. By examining the interplay between different pruning rates and optimizer performance, we provide detailed insights into the optimization dynamics that underpin StochGradAdam's superior results.


\begin{figure}[hbt!]
\centering
\includegraphics[width=1.0\columnwidth]{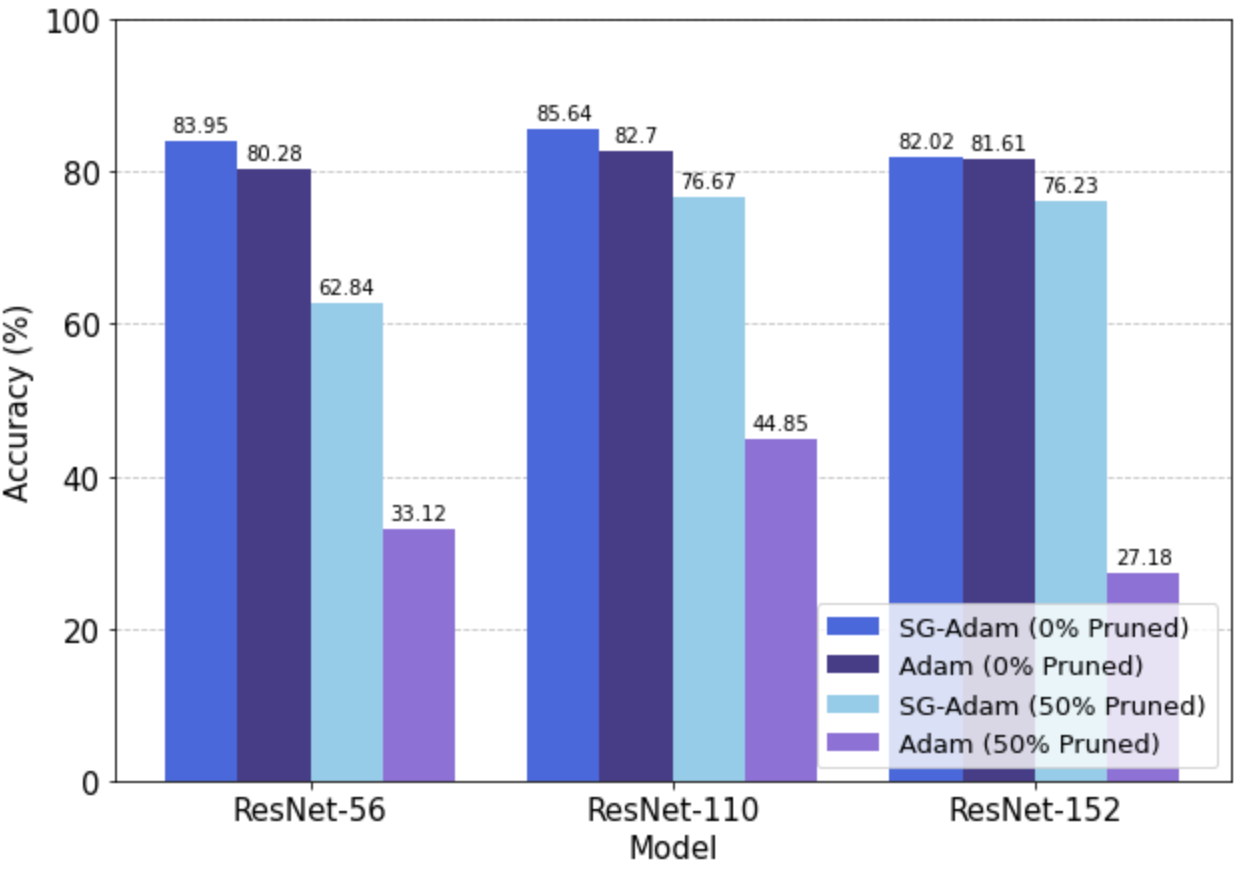}
\caption{Comparative Test Accuracy about Before and After 50\% Pruned ResNet Models~\cite{He2016ResNet} Trained with StochGradAdam~\cite{Yun2023StochGradAdam} (denoted as SG-Adam) and Adam Optimizers\cite{Kingma2014Adam} on CIFAR-10 Dataset\cite{Krizhevsky2009CIFAR10}}
\label{fig:eye}
\end{figure}

In summary, as detailed in the provided figure~\ref{fig:eye}, ResNet models~\cite{He2016ResNet} optimized with StochGradAdam~\cite{Yun2023StochGradAdam} consistently show higher accuracy than those optimized with Adam, both before and after applying a significant pruning rate. This trend holds across various model complexities, indicating a robust advantage in using StochGradAdam~\cite{Yun2023StochGradAdam} over Adam~\cite{Kingma2014Adam} for these network configurations. These results highlight the efficacy of StochGradAdam in optimizing neural networks prior to pruning, not merely in preserving a high degree of accuracy post-pruning but also in ensuring that the models remain adaptable and efficient for deployment in environments where computational resources are limited. 

\begin{figure*}[hbt!]
\centering
\includegraphics[width=2.0\columnwidth]{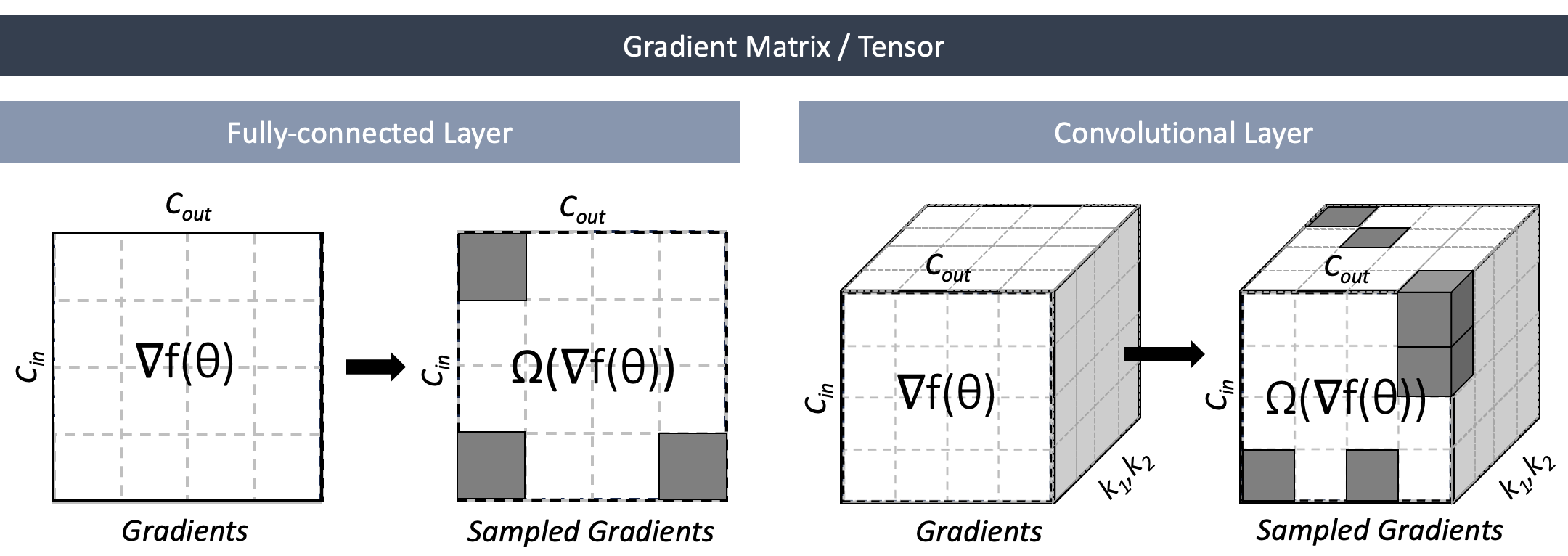}
\caption{Illustration of sampled gradient matrices/tensors for fully-connected and convolutional layers. On the left, the original gradients are presented with $C_{in}$ and $C_{out}$ denoting the number of input and output channels, respectively. On the right, the effect of gradient sampling is shown, where the application of a sampling function $\phi=\Omega(\nabla f(\theta))$ retains significant gradient components, represented by the white areas, and eliminates less significant ones, depicted as darkened blocks}
\label{fig:gs}
\end{figure*}

\section{Related Works}
The concept of neural network pruning, vital for reducing model complexity while retaining performance, has become increasingly relevant with the surge in deep learning applications~\cite{Seul2021}. Han et al. \cite{Han2015DeepCompression} pioneered the Deep Compression technique to significantly reduce model size without loss of accuracy. Liu et al.~\cite{Liu2017Learning} furthered this field with their network slimming approach that prunes channels in convolutional layers. Molchanov et al. \cite{Molchanov2019Importance} introduced a Taylor expansion-based criterion for this purpose, focusing on the importance of each weight in the loss function. Frankle and Carbin \cite{Frankle2019Lottery} introduced Lottery Ticket Hypothesis which shed light on the possibility of training pruned networks to match the accuracy of unpruned ones, provided the correct subset of weights is identified early. Li et al. \cite{Li2017PruningFilters} demonstrated substantial computational efficiency improvements with filter pruning. Yu et al.'s \cite{Yu2018NISP} neuron importance score propagation further elucidated the process of identifying and preserving important neurons in pruning.

Davis Blalock et al.~\cite{Blalock2020} provide a comprehensive review of neural network pruning and highlight that Magnitude-Based Pruning is one of the simplest yet effective methods to reduce computational complexity without severely impacting the model's accuracy. Song Han et al.\cite{Han2015LearningBW} demonstrate that learning both weights and connections for efficient neural networks through pruning can lead to significant reductions in storage requirements and improvements in computing efficiency. Pavlo Molchanov et al.\cite{Molchanov2019Importance} explore the importance estimation for neural network pruning and provide methods that improve upon simple Magnitude-Based Pruning, indicating the foundational role that magnitude-based methods play in the field. Zhuang Liu et al.\cite{Liu2018} delve into the value of network pruning and suggest that, in many cases, networks can be made significantly leaner without a drop in performance. Namhoon Lee et al.~\cite{Lee2020} provide a signal propagation perspective for pruning neural networks at initialization, discussing how early and simple pruning methods can predict the final performance of a network. 

The advent of advanced optimization techniques like Adam, developed by Kingma and Ba \cite{Kingma2014Adam}, and the exploration of gradient descent methods by Ruder~\cite{Ruder2016Overview} have revolutionized DNN training. Furthermore, StochGradAdam is introduced by Yun~\cite{Yun2023StochGradAdam}, which is using gradient sampling technique during optimization step. This optimization techique affect the distribution of the weights of trained model because it makes some gradients as zero. While Zhang et al. \cite{Zhang2018Systematic} have underscored the critical role of optimization strategies in the context of weight pruning, our methodology extends this concept by incorporating gradient sampling techniques, such as StochGradAdam \cite{Yun2023StochGradAdam}.


Magnitude-Based Pruning has been shown to effectively reduce the computational complexity~\cite{Han2015DeepCompression}, improve the speed and efficiency~\cite{Li2017PruningFilters}, and maintain or even enhance the generalization ability of neural networks~\cite{Molchanov2016PruningCN}. Its simplicity, coupled with its substantial impact on performance, makes it a widely adopted method in optimizing neural networks for various applications~\cite{Blalock2020}. We use this Magnitude-Based Pruning for post-prune the neural networks weights optimized with StochGradAdam~\cite{Yun2023StochGradAdam} which can affect the weight distribution during back-propagation step.

In essence, our study contributes to the evolving discussion on neural network pruning and optimization by proposing the innovative concept of gradient sampling optimization. This approach is designed to enhance the efficiency of neural networks and ensure accuracy retention, especially in resource-constrained environments. By combining gradient sampling techniques with pruning strategies, our methodology addresses a critical gap in neural network optimization, offering a valuable solution to the challenge of balancing efficiency and performance.

\section{Method}
This research investigates the efficacy of combining gradient sampling optimization, as implemented in StochGradAdam~\cite{Yun2023StochGradAdam}, with neural network pruning to maintain high model accuracy while reducing complexity. Our methodology comprises two primary stages: training the neural network using StochGradAdam and then applying pruning to the trained model.

\subsection{Training with StochGradAdam}
The initial stage of our methodology involves training the neural network using StochGradAdam~\cite{Yun2023StochGradAdam}, a variant of the Adam optimizer~\cite{Kingma2014Adam} enhanced with gradient sampling techniques. This approach is detailed in Algorithm \ref{algorithm:stochgrad}, where we outline the essential steps in implementing StochGradAdam.

\begin{algorithm}
\caption{StochGradAdam, a modified version of the Adam optimizer with gradient sampling.}
\begin{algorithmic}[1]
\Require Stepsize (Learning rate) $\alpha$
\Require Global Decay $\delta$
\Require Decay rates $\beta_1, \beta_2 \in [0, 1)$
\Require Initial parameter vector $\theta_0$
\Require sampling rate $s \in [0, 1)$
\Require Learning rate $\mu \geq 0$

\State Initialize $m$ and $v$ as zero tensors \Comment{Moment vectors}
\While{$\theta_t$ not converged}
    \State Get gradient $g$ with respect to parameters $\theta$
    \State Generate a random mask $\Omega$ from $\mathcal{U}(0,1)$
    \State $\Omega \leftarrow \begin{cases} 1 & \text{if } \Omega < s \\ 0 & \text{otherwise} \end{cases}$ 
    \State $\phi \leftarrow \Omega \circ g$ \Comment{Sampled gradient}
    \State $\beta_1^{t} \leftarrow \beta_1 \times \delta^t$
    \State $\beta_2^{t} \leftarrow \beta_2 \times \delta^t$
    \State $m_t \leftarrow \beta_1^{t} m_{t-1} + (1 - \beta_1^{t}) \phi$
    \State $v_t \leftarrow \beta_2^{t} v_{t-1} + (1 - \beta_2^{t}) \phi^2$
    \State $\hat{m_t} \leftarrow \frac{m_t}{1 - \beta_1^{t}}$
    \State $\hat{v_t} \leftarrow \frac{v_t}{1 - \beta_2^{t}}$
    \State $\theta \leftarrow \theta - \mu {m_{corr\_t}} / (\sqrt{v_{corr\_t}} + \epsilon )$ \Comment{Update parameters}
\EndWhile
\State \Return $\theta$ \Comment{Resulting parameters}
\end{algorithmic}
\label{algorithm:stochgrad}
\end{algorithm}

The StochGradAdam optimizer initializes with predefined settings such as the learning rate, beta coefficients for momentum and velocity, and a sampling rate that determines the proportion of gradients used in each update. Its core update formula is outlined as follows~\cite{Yun2023StochGradAdam}:
\begin{equation}
\theta_{t+1} = \theta_t - \mu \frac{\hat{m}_t}{\sqrt{\hat{v}_t} + \epsilon},
\end{equation}
wherein \( \alpha \) signifies the learning rate, \( \hat{m}_t \) is the bias-adjusted average of the gradients, and \( \hat{v}_t \) represents the bias-adjusted average of the squared gradients.

This optimizer also introduces a gradient sampling step. In this step, a mask generated from a uniform distribution is applied to each batch of gradients, effectively allowing for a selective inclusion of gradients. This selective process is essential for the adaptive characteristic of the optimizer. Given a gradient vector \( \mathbf{g} \), the optimizer's goal is to ascertain which components of \( \mathbf{g} \) will be utilized in the update. This selection is conducted using a stochastic mask \( \Omega \). Each element of \( \Omega \) is determined by a uniform random variable \( \mathcal{U}(0,1) \):
\begin{equation}
\Omega_t = 
\begin{cases} 
1 & \text{if } \mathcal{U}(0,1) < s, \\
0 & \text{otherwise},
\end{cases}
\end{equation}
\( s \) is the preset threshold that determines the fraction of gradients included for updates. After creating the stochastic mask \( \Omega \), the subsequent step is to calculate the sampled gradient \( \phi \) by performing an element-wise multiplication with the gradient vector \( \mathbf{g_t}\):
\begin{equation}
\phi_t = \Omega \odot \mathbf{g_t},
\end{equation}
where \( \odot \) symbolizes the element-wise multiplication. This operation ensures that only the gradient components flagged by the mask \( \Omega \) are factored into \( \phi \). This sampled gradient is shown as the figure~\ref{fig:gs}.

Moreover, the moving average of the gradients, indicated by \( m_t \), is updated at each iteration by an exponential decay method. This technique blends a portion of the previous moving average with the current sampled gradient:
\begin{equation}
m_t = \beta_1 m_{t-1} + (1 - \beta_1) \phi,
\end{equation}
where \( \beta_1 \) governs the rate of decay for the moving average. The adjusted decay rate at iteration \( t \), denoted by \( \beta_1^t \), is defined as $\beta_1^t = \beta_1 \odot \delta^t$ with \( \beta_1 \) balancing the influence of prior gradients—higher values weight the past more heavily, whereas lower values lean towards recent data.

The bias-corrected moving average \( m_t \) at the \( t^{th} \) iteration is given by:
\begin{equation}
\hat{m}_t = \frac{m_t}{1 - \beta_1^t},
\end{equation}
where \( 1 - \beta_1^t \) acts as a normalization constant to correct for the initial estimation bias.
Likewise, \( v \) denotes the moving average of the squared gradients. Its updating formula is articulated as:
\begin{equation}
v_t = \beta_2 v_{t-1} + (1 - \beta_2) \phi^2,
\end{equation}
where \( \beta_2 \) represents the exponential decay rate for the moving average of the squared gradients. Like \( \beta_1 \), \( \beta_2 \) is also time-adjusted, denoted by \( \beta_2^t \), and is calculated as $\beta_2^t = \beta_2 \odot \delta^t$. The operation \( \odot \) enables the element-wise squaring of each gradient component. The bias-corrected form of \( v \) at iteration \( t \) is:
\begin{equation}
\hat{v}_t = \frac{v_t}{1 - \beta_2^t},
\end{equation}
Both momentum (\( m \)) and velocity (\( v \)) are updated using the sampled gradients, integral to the adaptive learning nature of the optimizer. Governed by the beta parameters, these updates undergo bias correction to rectify the initial zero inclination. StochGradAdam then employs these corrected values to finely tune the network weights, ensuring a robust and efficient training pathway that preludes the subsequent phase of pruning.

\begin{figure*}[hbt!]
\centering
\includegraphics[width=2.0\columnwidth]{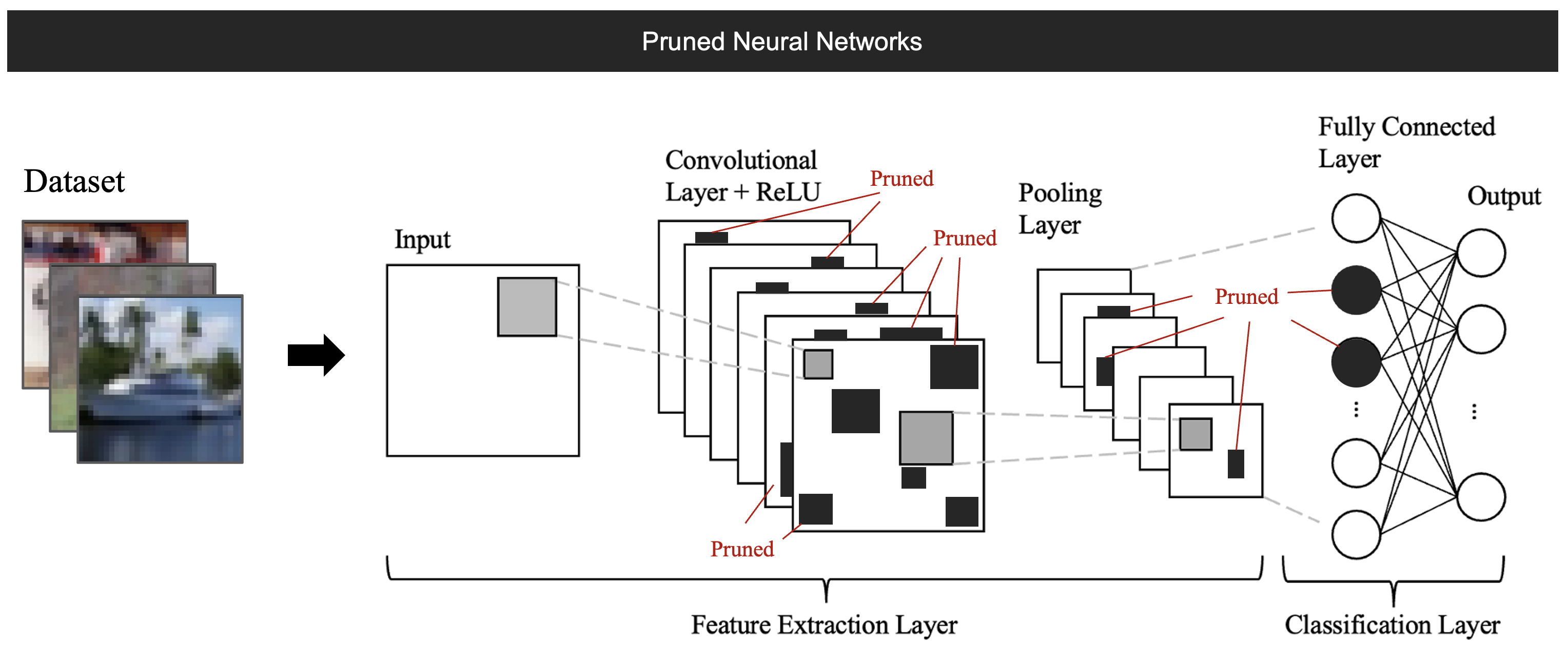}
\caption{This diagram demonstrates the application of Magnitude-Based Pruning with Percentile Threshold across all layers of a Convolutional Neural Network (CNN). It visualizes the process of pruning by setting weights to zero if they fall below a specific magnitude percentile, which is represented by the black areas in each layer.}
\label{fig:pruning}
\end{figure*}

\subsection{Pruning Process}
In our study, we explore a pruning technique to reduce the complexity of neural networks while maintaining high accuracy. The technique is Magnitude-Based Pruning, which targets weights with smaller magnitudes for removal~\cite{ko2019}. This method is integrated with the gradient sampling optimization used in StochGradAdam~\cite{Yun2023StochGradAdam}, allowing us to evaluate their effectiveness in preserving model accuracy post-pruning.
\vspace{0.2cm}

\textbf{Magnitude-Based Pruning.} Magnitude-based pruning approach is designed to reduce the complexity of neural networks by selectively removing weights based on their magnitudes \cite{Han2015LearningBW,Molchanov2016PruningCN, ko2019} as shown in~\ref{fig:pruning}. This method specifically targets weights at the lowest percentiles, under the assumption that these weights are less critical for the network's performance. This approach is detailed in Algorithm \ref{algorithm:pruning}~\cite{prun2024}.

\begin{algorithm}
\caption{Magnitude-Based Pruning with Percentile Threshold}
\begin{algorithmic}[1]
\Require Neural network with weights \( W \), Percentile \( P \)
\State Convert \( W \) to absolute values and sort in ascending order to get \( W_{\text{sorted}} \)
\State Calculate index \( idx = \left\lceil \frac{P}{100} \times |W| \right\rceil \) for the \( P^{th} \) percentile
\State Determine the pruning threshold \( \psi = W_{\text{sorted}}(idx) \)
\For{each weight \( w \) in \( W \)}
    \State Apply pruning: \( w' = 
        \begin{cases}
        0, & \text{if } |w| < \psi, \\
        w, & \text{otherwise},
        \end{cases} \)
\EndFor
\State \Return Network with pruned weights \( W' \)
\end{algorithmic}
\label{algorithm:pruning}
\end{algorithm}

To establish the pruning threshold \( \psi \), we utilize a statistical method by calculating a specific percentile within the distribution of the network’s weight magnitudes~\cite{prun2024}. Define \( W \) as the set containing the absolute values of all weights in the neural network. These weights are then arranged in ascending order to form \( W_{\text{sorted}} \). For a specified percentile \( P \), such as 40\%, \( \psi \) is identified at the position in \( W_{\text{sorted}} \) that corresponds to \( P\% \) of the length of \( W \). This method allows us to precisely define \( \psi \) based on the calculated percentile. Mathematically, it is calculated as:
\begin{equation}
\psi = W_{\text{sorted}}\left(\left\lceil \frac{P}{100} \times |W| \right\rceil\right),
\end{equation}
where \( |W| \) is the number of weights in \( W \), and \( \lceil \odot \rceil \) denotes the ceiling function, ensuring that the index is rounded up to the nearest integer. Applying the pruning rule, defined mathematically as:
\begin{equation}
w' = 
\begin{cases}
0, & \text{if } |w| < \psi, \\
w, & \text{otherwise},
\end{cases}
\end{equation}
where \( w \) denotes the individual weights of the neural network \cite{prun2024}. This approach effectively reduces the model size while aiming to retain the most significant weights, thereby maintaining the network's performance \cite{Han2015LearningBW,Molchanov2016PruningCN,prun2024}. The precise calculation of the pruning threshold and the systematic removal of the least significant weights ensure that the pruning process is both efficient and effective.

\section{Theoretical Analysis}
In this section, we analyze why models trained using the StochGradAdam optimizer maintain larger weight magnitudes and a higher variance in weights, contributing to their robustness against pruning compared to models trained with the traditional Adam optimizer. StochGradAdam employs a stochastic gradient sampling technique which allows it to skip some gradient updates. This method helps preserve regularization in the network, which are crucial for the model's performance, especially when subjected to weight pruning. \\

\textbf{Definition 4.1.} A function \(f: \mathbb{R}^d \rightarrow \mathbb{R}\) is convex if for all \(x, y \in \mathbb{R}^d\) and for all \(\lambda \in [0, 1]\)~\cite{Kingma2014Adam},
\begin{equation}
\lambda f(x) + (1-\lambda)f(y) \geq f(\lambda x + (1 - \lambda)y)
\end{equation}
Additionally, it is important to recognize that a convex function can be bounded from below by a tangent hyperplane at any point on the function. \\

\begin{figure*}[hbt!]
\centering
\includegraphics[width=2.0\columnwidth]{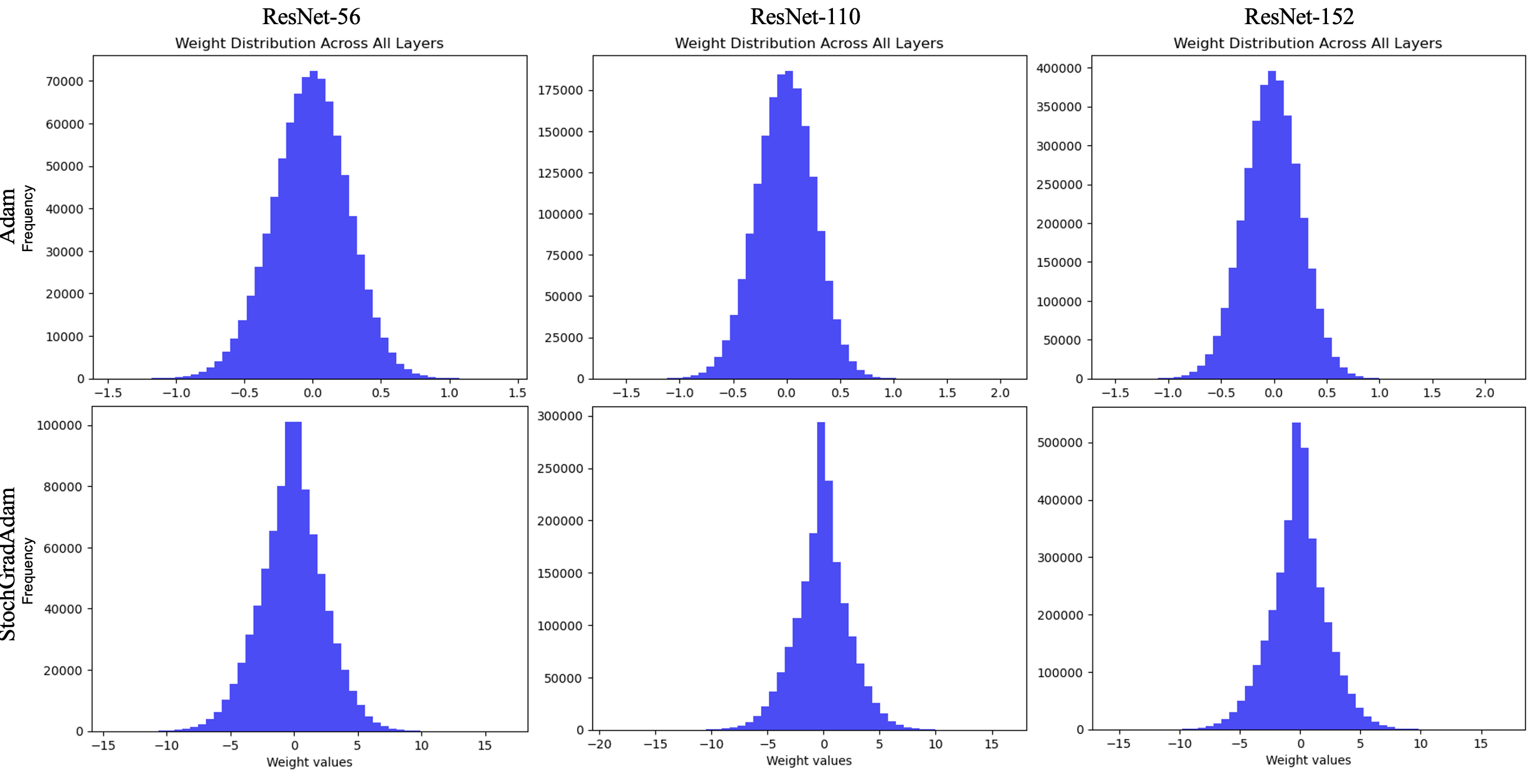}
\caption{Weight distributions of ResNet architectures trained with Adam and StochGradAdam optimizers. The top row shows histograms for ResNet-56, ResNet-110, and ResNet-152 trained with a standard Adam optimizer, exhibiting typical bell-shaped distributions centered around zero. The bottom histogram represents a ResNet model trained with the StochGradAdam optimizer, indicating a wider spread in weight values. }
\label{fig:hist}
\end{figure*}

\textbf{Lemma 4.2.} Let \( g_t = \nabla f_t(0) \), \( \phi_t = \Omega\nabla f_t(0) \), and \( g_{1:t} \), \( \phi_{1:t} \) be defined as above and bounded, \(||\phi||_{2} \leq\) \( ||g_t||_2 \leq G, ||g_{1:t}||_\infty \leq G_\infty \). This is reflected in the following inequality~\cite{Kingma2014Adam}:
\begin{align}
{\sum_{t=1}^{T} \sqrt{\frac{\phi_{t,i}^2}{t}}} 
\leq \sum_{t=1}^{T} \sqrt{\frac{g_{t,i}^2}{t}}
\leq 2G_\infty {||g_{1:T,i}||_2}
\end{align}
This inequality illustrates that the selected gradients through the StochGradAdam method contribute to an overall larger gradient norm across iterations, compared to the typical Adam method.\\

\textit{Proof.} We will establish the inequality through inductive reasoning on \( T \). Initially, for the base case when \( T = 1 \), it holds that $\sqrt{\phi_{1,i}^2} \leq \sqrt{g_{1,i}^2} \leq 2G_\infty {||g_{1,i}||_2}$. Continuing with the inductive step~\cite{Kingma2014Adam},
\begin{align}
\sum_{t=1}^{T} \sqrt{\frac{\phi_{t,i}^2}{t}} &= \sum_{t=1}^{T-1} \sqrt{\frac{\phi_{t,i}^2}{t}} + \sqrt{\frac{\phi_{T,i}^2}{T}} \\
 & \leq \sum_{t=1}^{T-1} \sqrt{\frac{g_{t,i}^2}{t}} + \sqrt{\frac{g_{T,i}^2}{T}} = \sum_{t=1}^{T} \sqrt{\frac{g_{t,i}^2}{t}} \\
 & \leq 2G_\infty||g_{1:T-1,i}||_2 + \sqrt{\frac{g_{i,T}^2}{T}} \\
 & = 2G_\infty\sqrt{||g_{1:T,i}||_2^2 - g_{T}^2} + \sqrt{\frac{g_{T,i}^2}{T}}
\end{align}

From $||g_{1:T,i}||_2^2-g^2_{T,i}+\frac{g^4{T,i}}{4||g_{1:T,i}||_2^2} \geq ||g_{1:T,i}||_2^2-g^2_{T,i}$, we can use square root for both side, and then we have, $\sqrt{||g_{1:T,i}||_2^2-g^2_{T,i}} \leq ||g_{1:T,i}||_2^2 - \frac{g^2_{T,i}}{2||g_{1:T,i}||_2^2} \leq ||g_{1:T,i}||_2^2 - \frac{g^2_{T,i}}{2\sqrt{TG^2_\infty}}$. Then, change the inequality and replace the term $\sqrt{||g_{1:T,i}||_2^2-g^2_{T,i}}$~\cite{Kingma2014Adam}: \vspace{-0.4cm}
\begin{align}
& G_\infty\sqrt{||g_{1:T,i}||_2^2 - g_{T}^2} + \sqrt{\frac{\phi_{T,i}^2}{T}} \\
\leq &G_\infty\sqrt{||g_{1:T,i}||_2^2 - g_{T}^2} + \sqrt{\frac{g_{T,i}^2}{T}} \vspace{0.5cm}\\
\leq &2G_\infty||g_{1:T,i}||_2
\end{align}

This lemma implies that the StochGradAdam method, by using $\phi_t,i$ instead of $g_t,i$, does not exceed the gradient accumulation that would be achieved by traditional gradient methods (like Adam), and it adheres to a specific bound. \\

\textbf{Theorem 4.3.} Let \( f : \mathbb{R}^d \rightarrow \mathbb{R} \) be a differentiable function optimized using gradients \(\nabla f(\theta_t) \), where \( \theta_t \in \mathbb{R}^d \) represents the parameters at iteration \( t \). Consider a mask \( \Omega \) where each \( \Omega_i \in \{0, 1\} \) for all \( i \), applied element-wise to the gradients. It holds that for each individual weight component \(i\),
\begin{equation}
|\theta_{t+1,i}^{\text{S}}| \geq |\theta_{t+1,i}^{\text{A}}|
\end{equation}
under the condition that \( \Omega \) selectively omits some gradient components. \\

\textit{Proof.} Assume \(\theta_{t,i}^A \leq \theta_{t,i}^S\) and $\nabla f(\theta_{{t,i}}^S) \leq \nabla f(\theta_{t,i}^A)$, we have for each component \(i\)~\cite{Goodfellow-et-al-2016}:
\begin{align}
(\theta_{t+1,i}^{\text{A}}) &= \theta_{t,i} - \eta \nabla f(\theta_{t,i}^A), \\
(\theta_{t+1,i}^{\text{S}}) &= \theta_{t,i} - \eta (\Omega_i \odot \nabla f(\theta_{t,i}^S)).
\end{align}

Given that \(\Omega_i\) is either 0 or 1 and selectively omits components, then \( |\Omega_i \nabla f(\theta_{t,i})| \leq |\nabla f(\theta_{t,i})| \). Thus, comparing the magnitude of updates for each component \(i\):
\begin{align}
|(\theta_{t+1, i}^{\text{S}})| &= | \theta_{t,i} - \eta (\Omega_i \nabla f(\theta_{t,i})) | \\
&\geq | \theta_{t,i} - \eta \nabla f(\theta_{t,i}) | \\
&= |(\theta_{t+1,i}^{\text{A}}) |.
\end{align}

Therefore, under the given conditions, the magnitude of each individual weight component in \( \theta_{t+1, i}^{\text{S}} \) is at least as large as that in \( \theta_{t+1, i}^{\text{A}} \), confirming that for each individual weight component, StochGradAdam does not decrease the magnitude more than Adam does, and may indeed increase it depending on the gradient components selected by \( \Omega \).

This selective update mechanism employed by StochGradAdam significantly influences the overall distribution of weights within the network. By selectively skipping some gradient updates, StochGradAdam effectively preserves larger weights, which enhances the model's robustness (more regularized), particularly against weight pruning processes. This preservation is crucial for maintaining important features in the model, ensuring that essential characteristics are not lost during optimization. 

\textbf{Weight Analysis.}
Figure~\ref{fig:hist} reveals the weight distributions for ResNet architectures when optimized with Adam and StochGradAdam across 200 epochs. The Adam optimizer tends to concentrate weights around zero, suggesting a traditional regularization effect~\cite{loshchilov2019decoupled, Santos_2022}. However, the distributions for models trained with StochGradAdam are not only more widely dispersed but also have a greater concentration of weights between 0 and 1 compared to those trained with Adam. This indicates a nuanced approach to regularization, where even though weights are allowed to explore a broader space, there is still a significant density of smaller magnitude weights. 

This denser presence of smaller weights in StochGradAdam-trained models implies more pronounced regularization, which is particularly beneficial for magnitude-based pruning techniques. Such pruning strategies rely on the assumption that weights with smaller magnitudes are less informative and therefore can be pruned with minimal loss to model performance~\cite{Han2015LearningBW,Molchanov2016PruningCN, ko2019}. StochGradAdam's tendency to retain a substantial number of small yet non-zero weights aligns well with the prerequisites of such pruning methods. The enhanced regularization and selective amplification of weights by StochGradAdam provide a dual benefit. On the one hand, the increased number of weights between 0 and 1 fosters robustness against overfitting by preventing the network from relying too heavily on a limited set of features. On the other hand, the preservation of important feature-representing weights ensures that the model's capacity for feature representation remains strong. This balance is essential for achieving effective generalization, making StochGradAdam an advantageous optimizer in the context of deep learning models aimed at both high accuracy and efficiency post-pruning.

\section{Experiments}
Prior to embarking on the pruning process, we conducted an initial training phase using the CIFAR-10 dataset~\cite{Krizhevsky2009CIFAR10} to evaluate the learning efficacy of two distinct optimizers, StochGradAdam~\cite{Yun2023StochGradAdam} and Adam~\cite{Kingma2014Adam}. CIFAR-10, a benchmark dataset comprising 60,000 32x32 color images across 10 classes, serves as an ideal proving ground for gauging the performance of neural network optimization due to its balanced variety of input features and complexity.

\vspace{0.5cm}
\textbf{Experimental Setting.}
In this critical phase of our research, we meticulously set up our experiments to evaluate the efficacy of StochGradAdam~\cite{Yun2023StochGradAdam} in neural network pruning, specifically focusing on its performance compared to the conventional Adam~\cite{Kingma2014Adam} optimizer. All experiments were conducted using an NVIDIA RTX 4080. We maintained a batch size of 128 across all experiments and We set the number of epochs as 200. Sampling rate for StochGradAdam is 0.8, and learning rate is 0.01 for both optimizers. Other hyperparameter settings such as $\beta_1=0.9, \beta_2=0.999, \epsilon=1E-7$ for both optimizers were kept at their default values and were identical for each.
\begin{table}[ht]
\caption{Test Accuracy of ResNet Architectures~\cite{He2016ResNet} on CIFAR-10 Dataset~\cite{Krizhevsky2009CIFAR10} Using StochGradAdam~\cite{Yun2023StochGradAdam} and Adam~\cite{Kingma2014Adam} Optimizers}
\vspace{0.25cm}
\small
\centering
\begin{tabular}{lcccc}
\hline
\hline
Optimizer & ResNet-56 & ResNet-110 & ResNet-152 \\
\hline
StochGradAdam & 83.95\% & 85.64\% & 82.02\% \\ 
Adam & 80.28\% & 82.70\% & 81.61\% \\ 
\hline
\hline
\end{tabular}
\label{tabmp}
\end{table}

\begin{figure*}[hbt!]
\centering
\includegraphics[width=2.0\columnwidth]{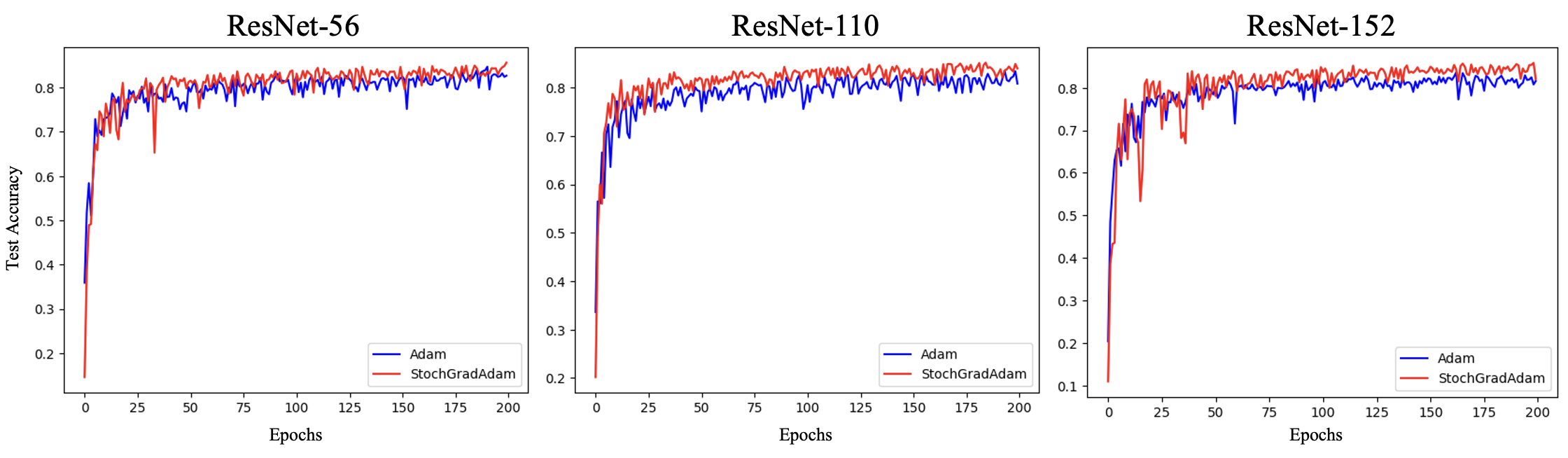}
\caption{Test Accuracy Comparison of StochGradAdam~\cite{Yun2023StochGradAdam} and Adam~\cite{Kingma2014Adam} Optimizers on ResNet Architectures~\cite{He2016ResNet} on CIFAR-10 Dataset~\cite{Krizhevsky2009CIFAR10}}
\label{fig:acc}
\end{figure*}

\begin{table*}\centering
\caption{Test Accuracy and Relative Accuracy Loss for ResNet Architectures~\cite{He2016ResNet} at Different Pruning Rates with CIFAR-10 Datasets~\cite{Kingma2014Adam}. \textbf{Bold} values indicate superior performance.}

\vspace{0.15cm}
\begin{adjustbox}{width=1\textwidth}
\scriptsize
 \renewcommand{\arraystretch}{1.3}
\begin{tabular}{@{}crrrrrrrrrrrrrrr@{}}\toprule

Pruning Rate & 
\multicolumn{4}{c}{ResNet-56} && 
\multicolumn{4}{c}{ResNet-110} && 
\multicolumn{4}{c}{ResNet-152}\\
\cmidrule{2-5} \cmidrule{7-10} \cmidrule{12-15}
(\%) 
& 
StochGrad & Adam & StochGrad & Adam && 
StochGrad & Adam & StochGrad & Adam && 
StochGrad & Adam & StochGrad & Adam \\
& 
Pruned & Pruned & Acc Loss & Acc Loss &&  
Pruned & Pruned & Acc Loss & Acc Loss &&
Pruned & Pruned & Acc Loss & Acc Loss \\
\midrule 

0\%  & \textbf{83.95\%} & 80.28\% & 0.00\% & 0.00\% && \textbf{85.64\%} & 82.70\% & 0.00\% & 0.00\% && \textbf{82.02\%} & 81.61\% & 0.00\% & 0.00\%\\
5\%  & \textbf{83.89\%} & 80.09\% & 0.07\% & 0.24\% && \textbf{85.67\%} & 82.63\% & +0.04\% & 0.08\% && \textbf{82.02\%} & 81.59\% & 0.00\% & 0.02\%\\
10\% & \textbf{83.60\%} & 81.23\% & 0.42\% & +1.18\% && \textbf{85.53\%} & 82.62\% & 0.13\% & 0.10\% && \textbf{82.01\%} & 81.73\% & 0.01\% & +0.15\%\\
15\% & \textbf{83.45\%} & 81.05\% & 0.60\% & +0.96\% && \textbf{85.36\%} & 82.02\% & 0.33\% & 0.82\% && \textbf{81.67\%} & 81.48\% & 0.43\% & 0.16\%\\
20\% & \textbf{82.96\%} & 80.22\% & 1.18\% & 0.07\% && \textbf{85.54\%} & 81.48\% & 0.12\% & 1.48\% && \textbf{82.87\%} & 82.30\% & +1.04\% & +0.85\%\\
25\% & \textbf{83.95\%} & 79.25\% & 0.0\% & 1.28\% && \textbf{85.17\%} & 80.83\% & 0.55\% & 2.26\% && \textbf{82.02\%} & 80.33\% & 0.00\% & 1.57\%\\
30\% & \textbf{81.31\%} & 78.32\% & 3.14\% & 2.44\% && \textbf{83.93\%} & 80.34\% & 2.00\% & 2.85\% && \textbf{81.97\%} & 78.99\% & 0.06\% & 3.21\%\\
35\% & \textbf{80.88\%} & 74.15\% & 3.66\% & 7.64\% && \textbf{82.84\%} & 78.17\% & 3.27\% & 5.48\% && \textbf{82.12\%} & 78.11\% & +0.12\% & 4.29\%\\
40\% & \textbf{76.95\%} & 70.82\% & 8.34\% & 11.78\% && \textbf{81.27\%} & 74.33\% & 5.10\% & 10.12\% && \textbf{80.70\%} & 74.88\% & 1.61\% & 8.25\%\\
45\% & \textbf{72.61\%} & 60.11\% & 13.51\% & 25.12\% && \textbf{79.72\%} & 61.93\% & 6.91\% & 25.11\% && \textbf{77.24\%} & 62.30\% & 5.83\% & 23.66\%\\
50\% & \textbf{62.84\%} & 33.12\% & 25.15\% & 58.74\% && \textbf{76.67\%} & 44.85\% & 10.47\% & 45.77\% && \textbf{76.23\%} & 54.67\% & 7.06\% & 33.01\%\\
55\% & \textbf{48.55\%} & 21.31\% & 42.17\% & 73.46\% && \textbf{68.08\%} & 19.49\% & 20.50\% & 76.43\% && \textbf{72.03\%} & 27.18\% & 12.18\% & 66.70\%\\
60\% & \textbf{32.10\%} & 10.80\% & 61.76\% & 86.55\% && \textbf{59.74\%} & 14.72\% & 30.24\% & 82.20\% && \textbf{63.15\%} & 11.90\% & 23.01\% & 85.42\%\\
65\% & \textbf{18.55\%} & 10.85\% & 77.90\% & 86.48\% && \textbf{42.68\%} & 11.93\% & 50.16\% & 85.57\% && \textbf{42.03\%} & 10.18\% & 48.76\% & 87.53\%\\
70\% & \textbf{11.64\%} & 10.08\% & 86.13\% & 87.44\% && \textbf{24.56\%} & 10.05\% & 71.32\% & 87.85\% && \textbf{21.01\%} & 11.24\% & 74.38\% & 86.23\%\\

\bottomrule
\end{tabular}
\end{adjustbox}
\label{tab:summary2}
\end{table*}
\textbf{Training Results.} In evaluating the optimization performance of StochGradAdam, our study focused on its deployment across different ResNet architectures during the training process. The outcomes, illustrated in figure \ref{fig:acc}, offer a clear comparative assessment of test accuracy between StochGradAdam and the conventional Adam optimizer across various ResNet models over 200 training epochs. We did not use pre-trained models or pre-trained weights for these experiments.

The analysis shows that StochGradAdam consistently outperforms Adam in terms of test accuracy, a pattern that holds true for ResNet-56, ResNet-110, and ResNet-152 models~\cite{He2016ResNet}. This enhancement in accuracy demonstrates StochGradAdam's effectiveness in navigating the intricate optimization challenges posed by deep neural networks.


\vspace{0.5cm}
\textbf{Pruning Results.} Following the comparative analysis (Table \ref{tab:summary2}) and noting the distinct performance advantage of StochGradAdam during the training phases, we expanded our investigation to include network pruning. Our results confirm that models trained with StochGradAdam not only deliver superior initial test accuracy but also sustain significantly better performance after pruning compared to those optimized with Adam.

When models undergo a 50\% reduction in parameters, the durability of StochGradAdam becomes especially apparent. For instance, the pruned ResNet-56 model achieves a test accuracy of 62.84\%, more than doubling the 33.12\% achieved by its counterpart trained with Adam. In the case of the more complex ResNet-110 and ResNet-152 architectures, StochGradAdam further asserts its superior optimization efficacy with post-pruning accuracies of 76.67\% and 76.23\% respectively, markedly outperforming the Adam optimizer, which yields accuracies of 44.85\% and 54.67\%. These outcomes clearly demonstrate that StochGradAdam enhances the robustness of neural networks, ensuring they retain high accuracy even when significantly pruned, thereby establishing a new standard for optimization techniques in the realm of model compression.



\begin{figure*}[hbt!]
\centering
\includegraphics[width=2.0\columnwidth]{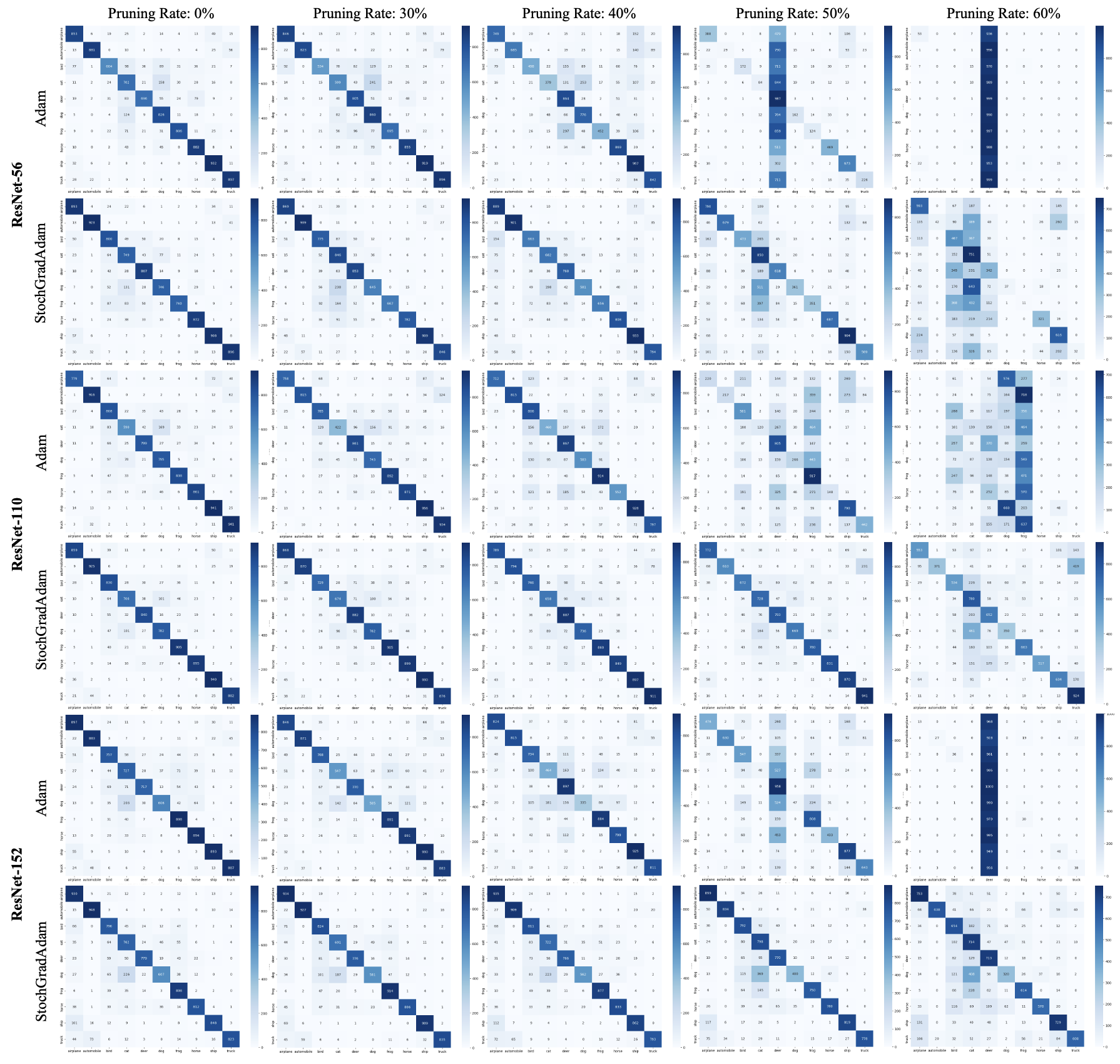}
\caption{Confusion matrices for pruned neural networks at various pruning rates (0\%, 30\%, 40\%, 50\%, and 60\%). Each row corresponds to a model utilizing a specific optimizer—Adam, StochGradAdam. The matrices visually represent the performance of each model in classifying different classes, with diagonal elements indicating correct predictions where the predicted class matches the actual class. This illustration can be used to compare the effects of pruning on model performance.}
\label{fig:conf}
\end{figure*}

Figure \ref{fig:conf} shows confusion matrices for pruned neural networks at various pruning rates (0\%, 30\%, 40\%, 50\%, and 60\%) correlate with our study's findings on the robustness of StochGradAdam optimization in the pruning process of neural networks. These matrices offer a visual assessment of each model's classification performance, utilizing optimizers such as Adam and StochGradAdam, with the diagonal elements representing correct predictions. It is evident from the matrices that as the pruning rate increases, models optimized with StochGradAdam maintain a higher density of correct classifications along the diagonal, compared to those optimized with Adam.

This observation aligns with the empirical results presented in our research, which highlight StochGradAdam's superior performance in maintaining accuracy during and after the pruning process. The matrices serve as a practical illustration of this performance difference. For instance, in the case of a 50\% pruning rate, the models optimized with StochGradAdam exhibit a significantly lesser decline in performance. This is characterized by a higher concentration of correct predictions in the confusion matrix's diagonal, indicating a robustness to pruning that is not as pronounced in models optimized with Adam.

The preservation of accuracy by StochGradAdam, as visually supported by the confusion matrices, offers insightful implications for neural network optimization strategies, particularly in resource-constrained environments where model compactness and efficiency are crucial. The ability of StochGradAdam to retain crucial information and maintain essential network functionalities post-pruning positions it as a promising optimization technique for developing efficient neural networks without substantial performance degradation.


\section{Discussion}


Our empirical results are not simply numerical victories; they signify a paradigm shift in the optimization strategies within the realm of model compression. StochGradAdam ensures that neural networks remain highly accurate even after aggressive pruning, challenging and potentially redefining the existing benchmarks for pruning methodologies.

The implications of this study are manifold. Firstly, it suggests that the adoption of advanced gradient sampling techniques can play a pivotal role in the future of neural network optimization, especially in the context of model compression. Secondly, it provides an empirical foundation for integrating sophisticated optimization techniques like StochGradAdam with pruning strategies, which can yield networks that are both efficient and highly accurate—a critical requirement in the age of ubiquitous and resource-constrained computing.

Moreover, these findings underscore the importance of considering optimizer choice as a fundamental aspect of network design, particularly when aiming for models that can maintain high accuracy in a compressed state. As we move forward, the insights garnered from StochGradAdam's performance could inform the development of new, even more, effective pruning strategies that further balance the trade-off between model size, computational efficiency, and accuracy retention.

\vspace{0.2cm}
\textbf{Limitation.} While our study has demonstrated the impressive capabilities of StochGradAdam in optimizing ResNet architectures, it's important to recognize the limitations of our experimental scope. The performance improvements attributed to StochGradAdam were primarily observed within ResNet models, which may inherently benefit more from the optimizer's gradient sampling techniques due to their unique design.

Focusing on ResNet models limits the broad applicability of our results. ResNets feature residual connections that help mitigate the vanishing gradient problem~\cite{He2016ResNet}, possibly enhancing the effectiveness of StochGradAdam's methodology~\cite{Yun2023StochGradAdam}. However, it's uncertain how well these benefits carry over to other architectures like densely connected networks, non-residual convolutional networks, or recurrent neural networks.

While our results confirm StochGradAdam's effectiveness with ResNet models, further studies are necessary to assess its performance across a wider range of neural network designs. Future research will expand our investigations to include diverse network types, offering a more holistic evaluation of StochGradAdam's utility. This broader analysis will help determine if the observed benefits are consistent across various architectures or specific to certain design features.

\section{Conclusion}
This study validates the effectiveness of the StochGradAdam optimizer, particularly within ResNet architectures, showing enhanced test accuracy and sustained performance even after significant parameter reductions. The selective gradient sampling approach of StochGradAdam is instrumental in these outcomes, especially notable in models subjected to 50\% pruning. Our experiments underscore StochGradAdam’s capability to significantly improve the robustness and accuracy of neural networks under pruning conditions. These promising results confirm StochGradAdam's potential within the specific context of ResNet architectures. Future research should extend this evaluation to various network designs to fully assess StochGradAdam's adaptability and effectiveness across different neural network architectures. This will help determine the broader applicability of StochGradAdam in enhancing neural network optimization in diverse machine learning applications.

\bibliographystyle{IEEEtran}
\bibliography{main}

\end{document}